\documentclass[conference]{IEEEtran}
\IEEEoverridecommandlockouts

\usepackage{cite}
\usepackage{amsmath,amssymb,amsfonts}
\usepackage{algorithmic}
\usepackage{graphicx}
\usepackage{textcomp}
\usepackage{xcolor}
\usepackage{booktabs}
\def\BibTeX{{\rm B\kern-.05em{\sc i\kern-.025em b}\kern-.08em
    T\kern-.1667em\lower.7ex\hbox{E}\kern-.125emX}}
\begin{document}

\title{Diagnosing Psychiatric Patients: Can Large Language and Machine Learning Models Perform Effectively in Emergency Cases?
}

\author{\IEEEauthorblockN{1\textsuperscript{st} Dr. Abu Shad Ahammed}
\IEEEauthorblockA{\textit{Chair of Embedded Systems} \\
\textit{University of Siegen}\\
Siegen, Germany \\
abu.ahammed@uni-siegen.de}
\and
\IEEEauthorblockN{2\textsuperscript{nd} Sayeri Mukherjee}
\IEEEauthorblockA{\textit{Chair of Embedded Systems} \\
\textit{University of Siegen}\\
Siegen,Germany \\
sayeri.mukherjee@student.uni-siegen.de}
\and
\IEEEauthorblockN{3\textsuperscript{rd} Prof. Dr. Roman Obermaisser} 
\IEEEauthorblockA{\textit{Chair of Embedded Systems} \\
\textit{University of Siegen}\\
Siegen, Germany \\
roman.obermaisser@uni-siegen.de}
}

\maketitle

\begin{abstract}
Mental disorders are clinically significant patterns of behavior that are associated with stress and/or impairment in social, occupational, or family activities. People suffering from such disorders are often misjudged and poorly diagnosed due to a lack of visible symptoms compared to other health complications. During emergency situations, identifying psychiatric issues is that's why challenging but highly required to save patients. In this paper, we have conducted research on how traditional machine learning and large language models (LLM) can assess these psychiatric patients based on their behavioral patterns to provide a diagnostic assessment. Data from emergency psychiatric patients were collected from a rescue station in Germany. Various machine learning models, including Llama 3.1, were used with rescue patient data to assess if the predictive capabilities of the models can serve as an efficient tool for identifying patients with unhealthy mental disorders, especially in rescue cases.
\end{abstract}


\section{Introduction}
In the current world, mental illness is a highly discussed topic, as symptoms can be invisible to other people, but the effect on patients can be severe. A significant number of people commit suicide due to the overburdening effect of mental illness. The multidimensional impact of severe mental illness includes physical health problems like intolerable headache, sleeplessness, lethargy, psychological difficulties, and socioeconomic drift \cite{fekadu2019multidimensional}. Among various diseases, the common mental illnesses found in rescue situations are depression, mania, alcoholic intoxication, insanity due to drugs, suicidal tendencies, high agitation, and psychosis. People with pre-existing mental health conditions can be severely affected by major accidents with a worsening impact on mental health, implying that rescuers should focus much of their attention on such patients and provide the necessary treatment \cite{rj1995psychiatric}.\\
Machine learning, a subset of artificial intelligence (AI), has become a strong transformative force in the healthcare industry, revolutionizing the way medical professionals diagnose health complications, predict treatment requirements, and optimize operational efficiencies in emergency care. In emergency medical situations, like rescue events, a patient's health assessment is not limited to the measurement of common health vitals. In complex situations where patients behave abnormally, rescuers must make a diagnosis from multiple perspectives and with foresight to determine whether the patient is in an exacerbating mental condition like panic attack, acute psychosis, or suicidal ideation. Time, being the main constraint in rescue situations, plays a vital role in decision making, as the rescuers are in a hurry to perform diagnostic tasks. AI-based solutions can be resourceful here due to their rapid decision making capability while initiating investigation into their potential reliability. Our novel research focuses on exploring this potential by integrating rescue patient data with machine learning and large language models. As health vitals alone cannot indicate such mental health situations, natural language processing (NLP) was used to analyze rescue cases to find important biomarkers before implementing in AI models. \\
The remainder of the manuscript is organized as follows. Section 2 provides a review of the relevant literature and a comprehensive background for our study. In Section 3, we detailed the research data and its characteristics. Section 4 describes the methods used for data organization and processing. Section 5 explores the application of NLP in psychiatric behavior analysis and how it is used to extract features, while Section 6 discusses the machine learning models deployed for analysis. In Section 7, we discuss on Llama model and how it is integrated for the psychiatric assessment. The results of the study are presented and discussed in Section 8, which includes insights based on different algorithms. Finally, Section 9 offers conclusions and potential directions for future research.
\section{Relevant Literature}
In our review of previous research, we found that although machine learning has been explored by scientists for the detection of psychiatric disorders, its use in helping emergency rescue patients in real time remains underexplored. The type of data used in our research to identify psychiatric patients, namely rescue data, is unprecedented in this research field, as it generally contains only the primary assessment of rescuers rather than medical notes from patients or doctors. Another research gap exists on the use of LLM like Llama on the diagnosis of emergency patients when controlled medical data is not available. There are some potential research \cite{zhao2024tuning, roy2025exploring, youssouf2025llamadrs, nowacki2025llm}, where the data source was clinical notes, social media posts, Patient Health Questionnaire (PHQ-9), or audio visual interviews with patients. Furthermore, the research was limited to specific mental illnesses such as depression, binary disorder, or anxiety, where our goal is to develop machine learning algorithms to understand the patient's mental state in general using limited raw rescue data. In most cases, patients were either not responsive or were unable to provide relevant information.\\
Previous research by Corcoran and Cecchi \cite{corcoran2020using} has explored the effectiveness of NLP and speech analysis in identifying psychosis and other psychiatric disorders. Their findings indicate that NLP can detect critical linguistic markers, including reduced semantic coherence and syntactic complexity, which are essential indicators of psychosis. In a related study, Cook et al. \cite{cook2016novel} demonstrated how combining NLP with machine learning algorithms significantly improves the early identification of individuals at risk of psychiatric symptoms.\\
In \cite{bzdok2018machine}, the authors discussed the opportunities and challenges in incorporating machine intelligence into psychiatric practice, as there is now evidence that machine learning can contribute to more accurate diagnoses and effective treatments than conventional diagnostic methods. Elujide et al. \cite{elujide2021application} in their article demonstrated that the integration of AI techniques can improve the early detection and classification of mental illness, for example, bipolar disorder, vascular dementia, attention deficit / hyperactivity disorder, insomnia and schizophrenia, offering a valuable tool for clinical decision support systems. Their deep learning model trained on the medical records of 500 patients with psychotic disorders achieved an accuracy of 75.17\% with an effective handling of class imbalance. However, there are also research works \cite{tate2020predicting, chen2020recent} that highlighted the difficulties and high misprediction rate of machine learning (ML) models in detecting mental illness due to data quality, model overfitting, or even lack of evidential features in the existing database.\\ 
\section{Research Data}
The initial raw rescue data was collected from the Siegen-Wittgenstein rescue station through a KMU-innovative research project named KIRETT. The data set comprises 273,183 unique cases from 2012 to 2021 and contains various health information from rescue patients. These records provide historical data on rescue operations for patients suffering from cardiovascular, respiratory, neurological, and psychiatric conditions, including depression and suicidal tendencies. First responders recorded each case, collecting rescue parameters, geographical details, patient medical history, clinical diagnoses, vital signs, initial impressions, medications administered, and treatment pathways. The data set was initially received in raw form, containing both structured and unstructured data, requiring data extraction and transformation for further analysis. A total of 452 attributes were identified as relevant for investigation. Both physiological and psychological evaluations were documented based on vital signs and observable behavioral and cognitive patterns, helping to detect the patient's health status. The rescue record was then systematically classified based on diagnosed complications, although some lacked diagnostic details due to patient reluctance, environmental restrictions, or fatal incidents. Despite these challenges, 10,220 rescue case data were successfully labeled as psychiatric cases.
\section{Data Organization and Preprocessing}
There were several challenges faced when working with the rescue data set, as the data was often incomplete, redundant, and did not provide meaningful information in raw form. This is not unexpected because during a rescue mission, it is quite challenging to collect data and make a diagnosis in a limited time frame. Hence, noisy artifacts, misspellings, and contaminated information were present in the data set that need to be filtered to extract meaningful information. In addition, the information was split into different files with duplicate values, which requires proper data organization and management before use for the training of the ML model. The following preprocessing steps were considered to make the raw data usable:
\begin{itemize}
    \item \textbf{Data Integration}: The database received through the KIRETT project consisted of 83 CSV files, each containing information about complete rescue events, while some records had multiple rows on the same patient to reflect how patient's health is evolving with time. Therefore, to further process the data, these files and patient information were merged using a Python program to obtain a holistic view of the patient's medical history, treatments, and results. The Python program went through each CSV file, identifying duplicate cases, and then merged them into a single row. If the information is the same in different records, then the particular cell contains only one data point. Otherwise, all different data are stored with a comma separator. After merging all these records using the case ID as key, the complete rescue database was saved in a XLSX file because of its user-friendly interface with better built-in tools for data analysis and presentation.
    \item \textbf{Data Reduction}: The rescue data set contained 452 columns, in which many of the rescue records were not usable as they did not contain significant information relevant to diagnosis. These records were identified and omitted for final data processing. In addition, there were some columns with duplicate information that were also deleted to shorten data processing times in the next steps.
    \item \textbf{Data Filtering}: As the raw data contained many outliers, the first step was to remove the definite ones, such as a negative value for a health vital (e.g. respiratory rate, pulse rate)  or ambiguous signs or expressions such as quotations and brackets in information columns that had no significance. The health vital data was later analyzed using the Interquartile Range (IQR) method via Python, with the aim of detecting additional outliers with values significantly deviating from the mean. These outliers were then substituted with their respective average values, as computed from the database. Correcting textual information was challenging and time consuming due to the size of the data and identifying whether the data should fall under a misspelled category or acronyms used due to a lack of time. Some of the empty cells where the values can be interpolated from other columns were filled with data imputation techniques.
\end{itemize}
After these basic steps, advanced data preprocessing was performed to identify relevant rescue cases for psychiatric patients by analyzing columns that have diagnostic information for patients. A total of 10,220 cases were found and selected that had adequate patient information. Cases of rescues where the patient could not be revived or where information was very limited have been excluded.
\section{Feature Engineering with NLP}
Feature engineering is highly significant in the machine learning pipeline and needs to be designed in a way so that all relevant attributes can be extracted, transformed, and modeled from the data set. However, extracting the right features for psychiatric complications posed certain challenges, as the vital signs are insignificant to identify a patient with mental illness. Thus, it was necessary for us to review the text columns that document information about rescue patients, including initial psychological evaluations, details of mental conditions such as indicators of depression, suicidal inclinations, or fear. To extract information from those columns, NLP, a subset of artificial intelligence, was used. NLP proves to be highly effective when the complexity of a set of textual data makes it challenging for humans to extract knowledge within a specific time frame \cite{chowdhary2020natural}. For an unstructured database like rescue records, NLP techniques play an important role in handling raw texts and extracting semantic meaning to understand the relevant context and intent. In our research, we used NLP to generate new features by parsing the text columns, identifying mainly repeated words, and checking if certain words related to psychiatric complications are present in the columns. The whole text parsing pipeline for an individual complication can be explained as follows:
\begin{itemize}
    \item Initially, all text columns relevant to the psychiatric complication of the rescue database were merged into a separate file. This was done for all relevant cases.
    \item A program called 'WordCounter' was written in Python that uses the text data file and checks how many times a word appears in that file. The Natural Language Tool Kit (NLTK) library was used to design this algorithm which covers both symbolic and statistical natural language processing, and is interfaced to annotated corpora \cite{bird2002}. It is one of the most popular platforms in the Python environment that allows users to access a wide range of functionalities through a consistent interface. Common tools and processes that come with NLTK are tokenization, parsing, classification, stemming, tagging, and semantic reasoning. \cite{farkiya2015natural}. The rescue text contains many repeated regular expressions and ambiguous words such as 'patient', 'since', and 'found', which need to be avoided. So, a stop-word dictionary was created that helped ignore those words during the word-counting process. Finally, the words most encountered (at least 50 times) were saved in an XLSX file and manually checked again to enlist the meaningful keywords.
    \item The initially selected keywords were then grouped into five different categories where the categories are nothing but certain health conditions common to a health complication. Each category is equivalent to a custom feature created from the text data. The five categories were: Preillness, alcoholism, psychiatric symptoms, abnormality, intoxication. In Table \ref{tab:psyfeature}, these categories and their relevant keywords are described.
    \item A Python-based algorithm was developed to parse all text data from rescue cases to identify a match for the specified keywords. If the text data of a rescue case contains one of the keywords, then its corresponding feature value was assigned to the matched keyword text in the feature column. It also meant that the feature was relevant for the rescue case. If no match was found, the feature value of the corresponding column was kept as none.
    \item The appearance of a certain keyword does not guarantee that the rescue case has relevance to that keyword, as negation words can also be present before the word in text columns to null out the possibility of any relevance. That is why the parsing algorithm was also included with the exceptions and assign keywords only if there is no negation before the word is present.
\end{itemize}
\begin{table*}[t]
\centering
\caption{Text Features For Psychiatry Complication}
\resizebox{\textwidth}{!}{
\begin{tabular}{|p{4cm}|p{6cm}|p{6cm}|}
\hline
\textbf{Feature} & \textbf{Description} & \textbf{Keywords} \\
\hline
Preillness & Any psychiatric pre-illness? & depression, mania, suicidality, agitation \\
\hline
Intoxication & Any medication overdose? & intoxication, LSD, drugs, cannabis, methadone, weed, pills \\
\hline
Alcoholism & Is the patient alcoholic? & drunk, alcohol, vodka, heroin, vodka (alt spelling), heavily intoxicated, ethanol (C2), cannabis, wasted \\
\hline
Mental Abnormalities & Any abnormal feelings or perceptions? & delusions of grandeur, fantasy, panic, panicked, fear, overdose, psych (abbr.), hallucination \\
\hline
Psychiatric Symptoms & Strong psychiatric symptoms observed? & no will to live, crying, devastated, suicidal, depressed, stress, confused, anxious, delusional, agitated, euphoric, aggressive, restless, withdrawal, borderline syndrome, borderline, excitation \\
\hline
\end{tabular}
}
\label{tab:psyfeature}
\end{table*}
Initially, selected features included data covering all accessible health measurements, with text-extracted features through NLP. But it required extra preprocessing to finalize features that can be used to develop the ML models. The primary objectives of the feature selection were to improve model performance, reduce overfitting, decrease training time, and improve model interpretability. In rescue modeling, it is not only about selecting the right features that are relevant to a health complication, but also removing features that are related to multiple complications. Because those common features will confuse the model during the training phase and will provide incorrect prediction results. In addition, eliminating unnecessary features will simplify the model, improve generalization capabilities, and reduce the computational burden associated with training and inference, as fewer features mean less complexity in the model \cite{nayak2023}.\\
Two popular filter selection methods, filter and wrapper, were applied in two stages of the final feature selection: pre-training and post-training. Before developing the ML model, the filter method is used to evaluate the relevance of features by their intrinsic properties, independently of any machine learning algorithm. An Excel-based validation algorithm was developed that evaluates the relevance score of the feature to estimate whether a keyword is more relevant to the specific complication than to other complications. To accomplish this, an average was calculated for each feature, representing the mean count of occurrences of that feature in both patients and non-patients for the psychiatric complication. Then a relative deviation was measured between these two averages to find the relevance score. The equation for relative deviation is as follows: 
\begin{equation}
\text{Relative Deviation} = \frac{|x - y|}{|y|} \times 100
\label{eq:relative_deviation}
\end{equation}
A small relative deviation indicates that the observed value is close to the reference value, while a large relative deviation indicates that the observed value has a high variance from the reference value. After observing these deviation values, for each individual complication, a threshold score of 3 was decided to finally select a feature. It indicated that the feature occurs at least three times more frequently in rescue events for a patient compared to non-patients or patients with different illnesses. After completing all the feature engineering steps, a total of ten features were chosen: five created using NLP and five related to health vitals, such as the Glasgow Coma Scale (GCS), blood circulation status (normal or abnormal), systolic blood pressure value, pulse rhythm status, and respiratory rate.
\section{Development of ML Models}
Like many other research fields, the implementation of machine learning in the health sector is increasing day by day. In emergency rescue scenarios, a well-known patient diagnostic approach, like CABCDE is used to determine the basic health status of the patient. Generally, these approaches are widely accepted by experts and likely improve the results by helping rescue personnel focus on the severe and life-threatening health complexities. However, there are certain limitations associated with these approaches when trying to diagnose the actual patient's health scenario as listed below:
\begin{itemize}
    \item Patients may not be able to communicate due to impaired consciousness. It causes a lack of access to the patient's medical history and prior diagnostic information
    \item Patients' representation of their symptoms and health complaints can vary widely and lead to false diagnosis
    \item Emergency patients may shows atypical symptoms due to the acute nature of their illness or injury which poses a risk of delayed or wrong diagnosis
    \item Due to high stress and time shortage, distortion of the cognitive abilities of first aid providers is quite common, potentially affecting their diagnostic accuracy
\end{itemize}
Taking into account these limitations, to improve patient care in rescue situations, machine learning algorithms are considered an effective solution due to their time-efficient predictive analysis capability. The ML algorithms chosen for predictive assessment are following:
\begin{itemize}
    \item Support vector machine (SVM)
    \item Random forest (RF)
    \item Extreme gradient boosting (XGB)
    \item K-nearest neighbors (K-NN)
    \item Naive Bayes (NB)
    \item Logistic regression (LR)
    \item Multilayer perceptron classifier (MLPC)
\end{itemize}
Most of the models selected above are classical algorithms, as they have demonstrated substantial potential to enhance diagnostic accuracy and address complex health issues, particularly when dealing with structured and unstructured data that may be limited or noisy \cite{rajkomar2019machine}.\\
The first step in the model development pipeline was to prepare the training data with the selected features. As these data will guide the machine learning algorithm on the pattern of both patients and non-patients, it should contain data not only from the specific health complications, but also from patients with other complications and healthy persons to maintain data balance. Upon examining non-patient cases, 8,758 instances were identified that could be included in the training dataset, given the availability of chosen feature values.\\
For tuning, we adapted both grid search and random search approach. In grid search, it runs through a predefined subset of the hyperparameter space in all feasible settings to identify the combination that yields the best model performance based on the chosen evaluation criteria. Random search is also a popular method for hyperparameter tuning, where a random combination of hyperparameter values is selected from a predefined range or distribution. For both tuning methods, hyperparameters were selected with a range of possible values that greatly contribute to the model performance. The evaluation metric 'Accuracy' was chosen to determine the performance of the ML model with a subset of hyperparameter values defined in the grid. To ensure that the model performance is robust and not dependent on a particular data split, StratifiedKfold and Kfold cross-validation were implemented. After the tuning was completed, the selected hyperparameters were then used to run the selected machine learning models. The goal was to find the algorithm that provides the best results in terms of evaluation criteria. Not all features will significantly contribute to the model evaluation, so we applied recursive feature elimination with cross-validation (RFECV), an advanced feature selection technique to select the best subset of features in terms of model performance.
\section{Integration of Llama}
Llama, an open source member of the trending large language models (LLM), has shown high performance on the natural language processing benchmark, making it a significant tool for health care analysis\cite{adams2024llama}. Released first in April 2024, it offers three different sizes with 8B, 70B, and 405B parameters. In our research, we have used Llama 8B model, as its incorporation can significantly improve rescue care and provide an explanation for understanding the reasoning behind its recommendations. However, these models can generate inaccurate responses \cite{boggavarapu2024evaluating} and can be fatal if used solely in psychiatric diagnosis because unlike physical health metrics, psychiatric symptoms are described in ambiguous natural language, vary between individuals, and are highly context dependent. Also, large language models like Llama are trained on large-scale text corpora from different resources containing misinformation, biases, and ambiguity \cite{fer2023, chen2024}. That's why our focus was to investigate with limited test case data to check the effectiveness of Llama in the diagnosis of psychiatric patients if task-specific test data are provided.\\
To use the Llama model, we installed Ollama that can run, manage, and interact with LLM and can handle data locally without cloud dependency. Afterwards, a Python program was written to run the specific Llama model using Ollama. Two variables were created: message and response, which represented prompt and response on patient diagnosis, respectively. Six test data were randomly chosen with 3 patients and 3 non-patients as originally diagnosed. The selected test data were then modified to a suitable format so that Llama could easily understand the context. The approach we implemented is known as zero-shot learning, where the model is not given any examples of the task at hand. It needs to infer the task solely based on the prompt (instructions) and its prior knowledge learned during pre-training. A sample of the prompt based on test data is provided below:
\vspace{10pt}
\\
'Systolic Blood Pressure': 170,\\
'Respiratory Rate': 13,\\
'Blood Circulation Normality': 1,\\
'GCS': 15,\\
'Pulse Rhythm': False,\\
'Any Preillness': False,\\
'Mental Sickness Possibility': False,\\
'Psychiatric Syndrom Presence': False,\\
'Alcoholic Possibility': False,\\
'Intoxication Possibility': False\\
\textit{Based on the above data collected from patient, please reply with true or false if the patient can be diagnosed as psychiatric patient}
\section{Results and Discussion}
\begin{figure}[]
    \centering
    \includegraphics[width=1\linewidth]{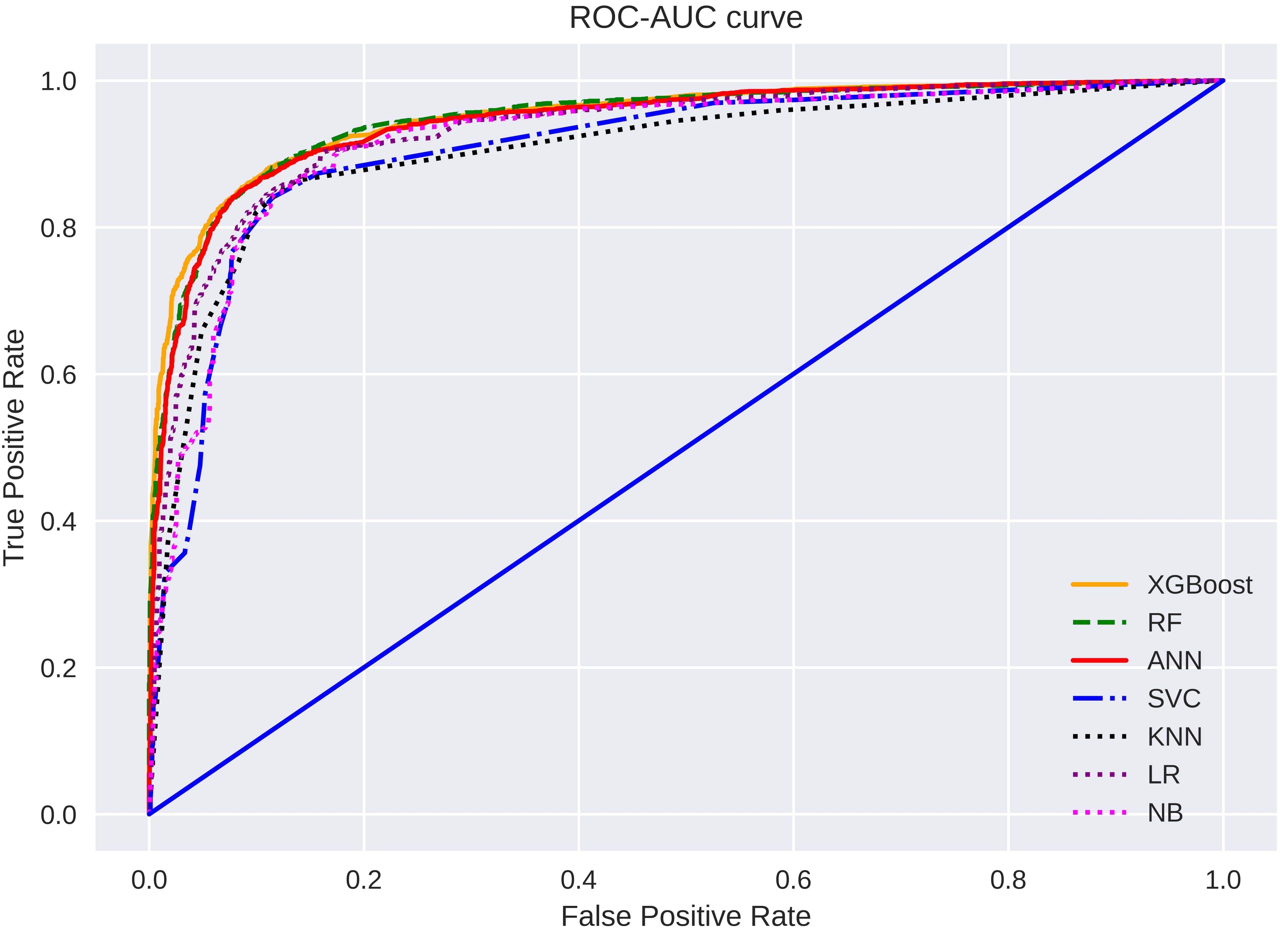}
    \caption{ROC-AUC Curve to Depict Performance of Psychiatric Complication Detection Algorithms}
    \label{fig:psyroc}
\end{figure}

The performance of the ML models was evaluated using several standard evaluation metrics to get a multidimensional view of the strengths and weaknesses of the model. Initially, we splitted the training data into two parts: train and test with a ratio of 80\% - 20\% and then use these data to the trained ML model to check if it correctly identifies patient and non-patient. Based on it's prediction, we created a confusion matrix showing the counts of true positives (TP), false positives (FP), true negatives (TN), and false negatives (FN) when compared with the original data. These components helped to evaluate performance metrics: accuracy, precision, specificity, sensitivity, and F1 score.\\
As can be seen in Table \ref{table:psyperf}, it reveals that RF had the best performance among other algorithms. It achieved an accuracy of 89.27\% and the highest F1 score of 90.01\%, indicating a strong balance between precision and recall. XGBoost followed closely with slightly lower values, demonstrating strong predictive capability with balanced sensitivity and specificity. The ANN model achieved high specificity (90.58\%) and precision (91.54\%), which highlights its strength in correctly identifying negative cases and its high confidence in positive predictions, although its recall is slightly lower than that of Random Forest. Naive Bayes, Logistic Regression and SVM had slightly lower scores, with Naive Bayes showing decent sensitivity (83.99\%) and high precision (90.11\%), suggesting that it is relatively reliable in identifying positive cases but not as balanced as Random Forest.
\begin{table}[!b]
\centering
\small
\caption{Performance metrics of ML models for psychiatric complication detection (values in \%)}
\renewcommand{\arraystretch}{1.1}
\setlength{\tabcolsep}{4pt}
\begin{tabular}{|l|c|c|c|c|c|}
\hline
\textbf{Model} & \textbf{Accuracy} & \textbf{Sensitivity} & \textbf{Specificity} & \textbf{Precision} & \textbf{F1-score} \\
\hline
RF & \textbf{89.27} & \textbf{89.44} & 89.07 & 90.59 & \textbf{90.01} \\
\hline
XGB & 87.97 & 87.17 & 88.91 & 90.25 & 88.68 \\
\hline
MLPC & 87.90 & 85.65 & \textbf{90.58} & \textbf{91.54} & 88.49 \\
\hline
NB & 86.30 & 83.99 & 89.05 & 90.11 & 86.94 \\
\hline
LR & 86.06 & 83.74 & 88.92 & 90.34 & 86.91 \\
\hline
SVM & 85.88 & 86.81 & 84.81 & 86.81 & 86.81 \\
\hline
K-NN & 85.74 & 82.98 & 89.00 & 89.92 & 86.31 \\
\hline
\end{tabular}
\label{table:psyperf}
\end{table}
The ROC-AUC curve for the psychiatric complication detection module is shown in Figure \ref{fig:psyroc}. Based on this figure, XGBoost, Random Forest and Artificial Neural Network algorithms demonstrated the highest ROC values while closely touching the upper left corner. This indicates a strong balance between the sensitivity and specificity of these models. The models showed a high area under the curve values, which meant that they were highly effective in both detecting true positives and avoiding false positives.
\\
All features selected previously, except for Preillness, were found valuable by RFECV to build the model that represents the above performance. So we skipped it when designing the Llama prompt. The response of Llama to selected test data from rescue patients can be found in Table \ref{tab:perf}:
\begin{table}[!h]
\centering
\caption{Llama Prediction on Test Cases. Prediction Result True Means the Patient was Recognized as Mentally Ill and Vice Versa}
\setlength{\tabcolsep}{3pt} 
\renewcommand{\arraystretch}{1.1} 
\begin{tabular}{|l|c|c|c|c|c|c|}
\hline
\textbf{Feature} & \textbf{Test1} & \textbf{Test2} & \textbf{Test3} & \textbf{Test4} & \textbf{Test5} & \textbf{Test6} \\
\hline
Sys. BP & 130 & 100 & 142 & 158 & 130 & 180 \\ \hline
Respiratory Rate & 16 & 14 & 15 & 12 & 16 & 16 \\ \hline
Circulation & Normal & Normal & 0 & 3 & 3 & 0 \\ \hline
GCS & 12 & 15 & 15 & 12 & 15 & 14 \\ \hline
Pulse Rhythm & FALSE & FALSE & FALSE & FALSE & FALSE & FALSE \\ \hline
Mental Abnormality & FALSE & FALSE & FALSE & FALSE & TRUE & FALSE \\ \hline
Psy. Syndrom & FALSE & FALSE & TRUE & FALSE & TRUE & FALSE \\ \hline
Alcoholic & TRUE & FALSE & FALSE & FALSE & FALSE & FALSE \\ \hline
Intoxication & TRUE & FALSE & FALSE & FALSE & FALSE & FALSE \\ \hline
\multicolumn{7}{|c|}{\textbf{--- Prediction Results ---}} \\ \hline
\textbf{ML Prediction} & \textcolor{red}{TRUE} & FALSE & TRUE & FALSE & TRUE & FALSE \\ \hline
\textbf{Llama Prediction} & \textcolor{red}{FALSE} & FALSE & TRUE & FALSE & TRUE & FALSE \\ \hline
\end{tabular}

\label{tab:perf}
\end{table}\\
For test cases, the original rescuer diagnostic was similar to the prediction of the ML model. Also, the prediction result from Llama was not too far from the best-performing ML model or the original diagnostic, since only one mismatch occurred. This indicates that with zero-shot learning, Llama has the potential to provide a reliable prediction for psychiatric evaluation if the prompt is designed efficiently.
\section{Conclusion}
The machine learning models developed in this study demonstrated high precision, underscoring the potential of integrating advanced AI technologies into mental health care. With the integration of Llama model we found reliable prediction performance with limited test case data. Future work will focus on expanding the test dataset and assessing the reliability of different LLM models for the diagnosis of psychiatric patients in the broad spectrum.
\bibliographystyle{IEEEtran}
\bibliography{main}

\end{document}